\documentclass[twoside,11pt]{article}

\usepackage{jmlr2e}

\usepackage{todonotes}
\usepackage{hyperref}
\usepackage{overpic}

\newcommand{\code}[1]{\texttt{#1}}

\jmlrheading{1}{2023}{1-48}{4/00}{10/00}{Antonio Carta, Lorenzo Pellegrini, Andrea Cossu, Hamed Hemati, Vincenzo Lomonaco}

\ShortHeadings{Avalanche}{Carta, Pellegrini, Cossu, Hemati, Lomonaco}
\firstpageno{1}

\begin{document}

\title{Avalanche: A PyTorch Library for Deep Continual Learning}

\author{\name Antonio Carta \email antonio.carta@unipi.it \\
       \addr University of Pisa
       \AND
       \name Lorenzo Pellegrini \email l.pellegrini@unibo.it \\
       \addr University of Bologna
       \AND
       \name Andrea Cossu \email andrea.cossu@sns.it \\
       \addr Scuola Normale Superiore
       \AND
       \name Hamed Hemati \email hamed.hemati@unisg.ch \\
       \addr University of St. Gallen
       \AND
       \name Vincenzo Lomonaco \email vincenzo.lomonaco@unipi.it \\
       \addr University of Pisa
       \AND}

\editor{-}

\maketitle

\begin{abstract}
Continual learning is the problem of learning from a nonstationary stream of data, a fundamental issue for sustainable and efficient training of deep neural networks over time. Unfortunately, deep learning libraries only provide primitives for offline training, assuming that model's architecture and data are fixed. Avalanche  is an open source library maintained by the ContinualAI non-profit organization that extends PyTorch by providing first-class support for dynamic architectures, streams of datasets, and incremental training and evaluation methods. Avalanche provides a large set of predefined benchmarks and training algorithms and it is easy to extend and modular while supporting a wide range of continual learning scenarios. Documentation is available at \url{https://avalanche.continualai.org}.
\end{abstract}

\begin{keywords}
  Continual Learning, lifelong learning, PyTorch, reproducibility
\end{keywords}

\section{Introduction}

Learning continually from non-stationary data streams is a long-sought goal in Artificial Intelligence. While most deep learning methods are trained offline, there is a growing interest in Deep Continual Learning (CL)~\citep{lesortContinualLearningRobotics2020} to improve learning efficiency, robustness and adaptability of deep networks. Deep learning libraries such as PyTorch and Tensorflow are designed to support offline training, making it difficult to implement continual learning methods.
Avalanche\footnote{Official Avalanche website: https://avalanche.continualai.org}, initially proposed in~\cite{lomonacoAvalancheEndtoEndLibrary2021}, provides a comprehensive library to support the development of research-oriented continual learning methods. The library is maintained by the \emph{ContinualAI} non-profit organization. Compared to existing continual learning libraries~\citep{douillardContinuumSimpleManagement2021,wolczykContinualWorldRobotic2021,normandinSequoiaSoftwareFramework2022,mirzadehCLGymFullFeaturedPyTorch2021,masanaClassIncrementalLearningSurvey2022}, Avalanche supports a larger number of methods and scenarios. Avalanche is a library with a growing user base\footnote{https://github.com/ContinualAI/avalanche}, an increasing set of features, and a strong focus on reproducibility(\href{https://github.com/continualAI/continual-learning-baselines}{link}).

\section{What can you do with Avalanche?}
\begin{figure}[t]
    \centering
    \includegraphics[width=0.99\textwidth]{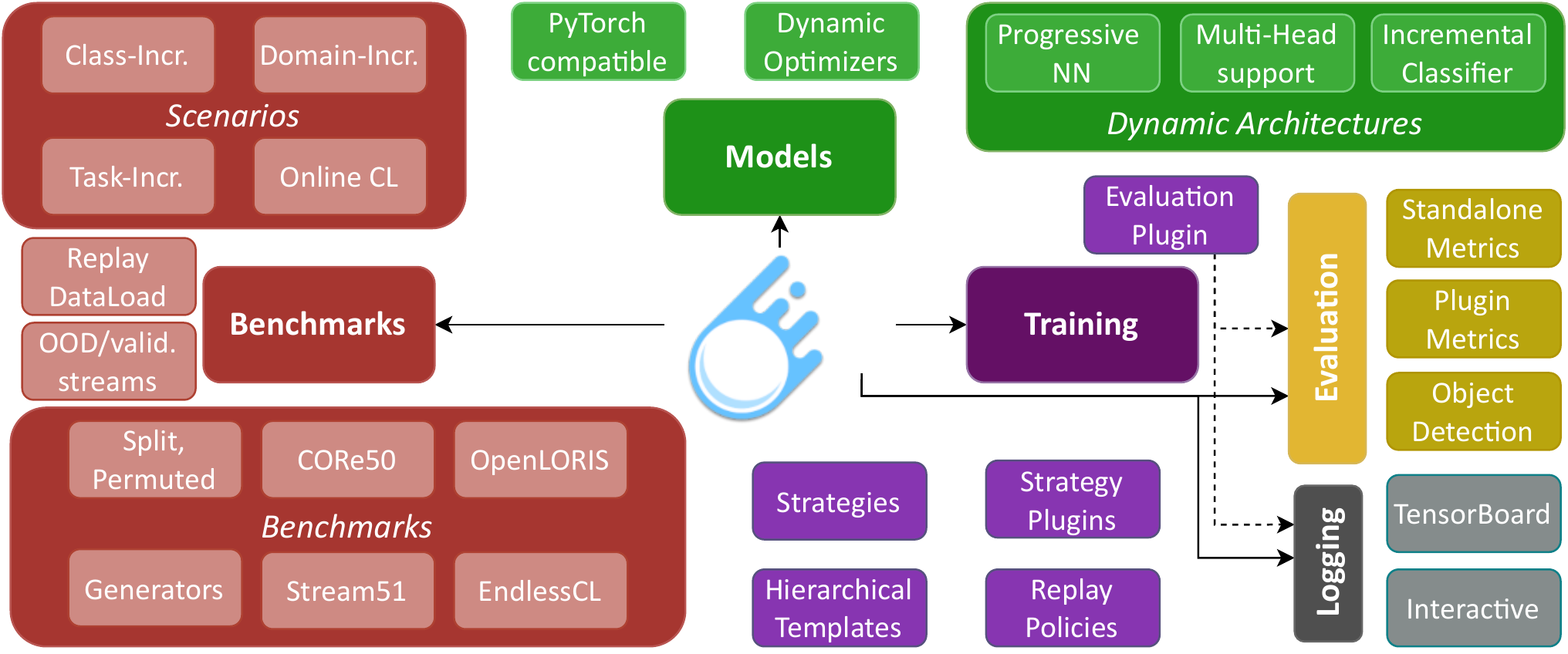}
    \caption{Avalanche main functionalities and modules.}
    \label{fig:modules}
\end{figure}

Avalanche is a library built on top of PyTorch~\citep{paszkePyTorchImperativeStyle2019}, designed to provide simple and stable components with everything that you need to execute continual learning experiments. The library is split into 5 modules: benchmarks, training, models, evaluation, and loggers. Figure \ref{fig:modules} shows a high-level overview of the library and its major components.

\paragraph{Benchmarks} provides standard benchmark definitions (\href{https://avalanche-api.continualai.org/en/v0.3.1/benchmarks.html#classic-benchmarks}{doc}), high-level benchmark generators (\href{https://avalanche-api.continualai.org/en/v0.3.1/benchmarks.html#benchmark-generators}{doc}), and low-level utilities to define new benchmarks by manipulating stream of experiences and datasets (\href{https://avalanche-api.continualai.org/en/v0.3.1/benchmarks.html#utils-data-loading-and-avalanchedataset}{doc}).

\paragraph{Training} provides standard training algorithms. Training strategies can be easily extended and combined with each other. Many continual learning techniques are available to be used out-of-the-box (\href{https://avalanche-api.continualai.org/en/v0.3.1/training.html#training-strategies}{doc}). For replay methods, custom storage policies can be implemented, and many options such as balancing methods and reservoir sampling are available (\href{https://avalanche-api.continualai.org/en/v0.3.1/training.html#replay-buffers-and-selection-strategies}{doc}). Most training methods can be combined together to create hybrid strategies.

\paragraph{Models} provides CL architectures and first-class support for dynamic architectures, multi-task models, and optimizer update (\href{https://avalanche-api.continualai.org/en/v0.3.1/models.html#}{doc}). \code{DynamicModule}s implement growing architectures such as multi-head classifiers and progressive networks~\citep{rusuProgressiveNeuralNetworks2016}.

\paragraph{Evaluation} provides CL metrics (\href{https://avalanche-api.continualai.org/en/v0.3.1/evaluation.html}{doc}). A set of metrics is declared during the strategy initialization and is automatically computed during the training and evaluation loops. Avalanche provides an extensive set of metrics to measure the model's performance and to keep track of system's metrics such as memory occupation and CPU usage. All the metrics can be computed at different granularities (e.g. minibatch, experience, stream).

\paragraph{Logging} metrics are collected and serialized automatically by the logging system. The \code{EvaluationPlugin} (\href{https://avalanche-api.continualai.org/en/v0.3.1/generated/avalanche.training.plugins.EvaluationPlugin.html}{doc}) connects training strategies, metrics, and loggers, by collecting all the metrics and dispatching them to all the registered loggers. TensorBoard, Weights and Biases, CSV files, text files, and standard output loggers are available (\href{https://avalanche-api.continualai.org/en/v0.3.1/logging.html}{doc}), but the logging interface can be easily extended with new loggers for custom needs.

\paragraph{Core Utilities} Avalanche offers a checkpointing functionality, allowing to pause and resume experiments. All Avalanche components are serializable.

\section{API and Design}
At the high level, Avalanche provides ready-to-use strategies, which can be instantiated and trained with a minimal amount of code (\href{https://github.com/ContinualAI/avalanche/blob/master/README.md}{example}). Internally, the training and eval loops implement a callback system that supports external plugins. Plugins allow to build on top of existing strategies with minimal changes and to combine different strategies together, a critical feature that other CL libraries do not support. Thanks to the plugin system, many strategies can be easily reused in different scenarios or combined together without any change to their code.

\paragraph{Benchmarks, Streams, Experiences} Benchmarks in Avalanche provide the data needed to train and evaluate CL models. \emph{Benchmarks} are a collection of streams (e.g., a train and test stream for SplitMNIST~\citep{lomonacoAvalancheEndtoEndLibrary2021}). \emph{Streams} are sequences of \emph{Experiences}, where an experience stores all the information available at a certain point in time. Experiences provide all the necessary information for training, evaluation and logging. For example, in supervised CL experiences provide a dataset, while in reinforcement learning they provide an environment. Streams and experiences have private unique identifiers for logging purposes.

\paragraph{Dataset and Stream Manipulation} \code{AvalancheDataset} extends PyTorch datasets with the ability to add attributes values (such as task labels and other metadata) at the example granularity, manage different groups of transformations, and define custom collate functions. \code{AvalancheDataset}s can be subsampled and concatenated, providing a simple API to define and manipulate streams and replay buffers. In addition, Avalanche offers flexible data loaders that control balancing and joint sampling from multiple datasets.

\paragraph{Dynamic and Standalone Components} Avalanche extends many PyTorch static components into dynamical objects. For example, CL strategies may require changing the model's architecture, optimizer, losses, and datasets during training. In PyTorch, these are static objects that are not easy to update during training (e.g., the architecture of nn.Modules is fixed). Avalanche \code{DynamicModule}s provide a simple API to update the model's architecture, \code{ExemplarsBuffer} manage replay buffers, regularization plugins update the loss function after each experience, and optimizer are also updated before each experience. Every component is automatically managed by Avalanche strategies, but they can also be used standalone in a custom training loop (\href{https://avalanche.continualai.org/how-tos/dataloading_buffers_replay}{example}).

\begin{figure}
    \centering
    \includegraphics[width=0.99\textwidth]{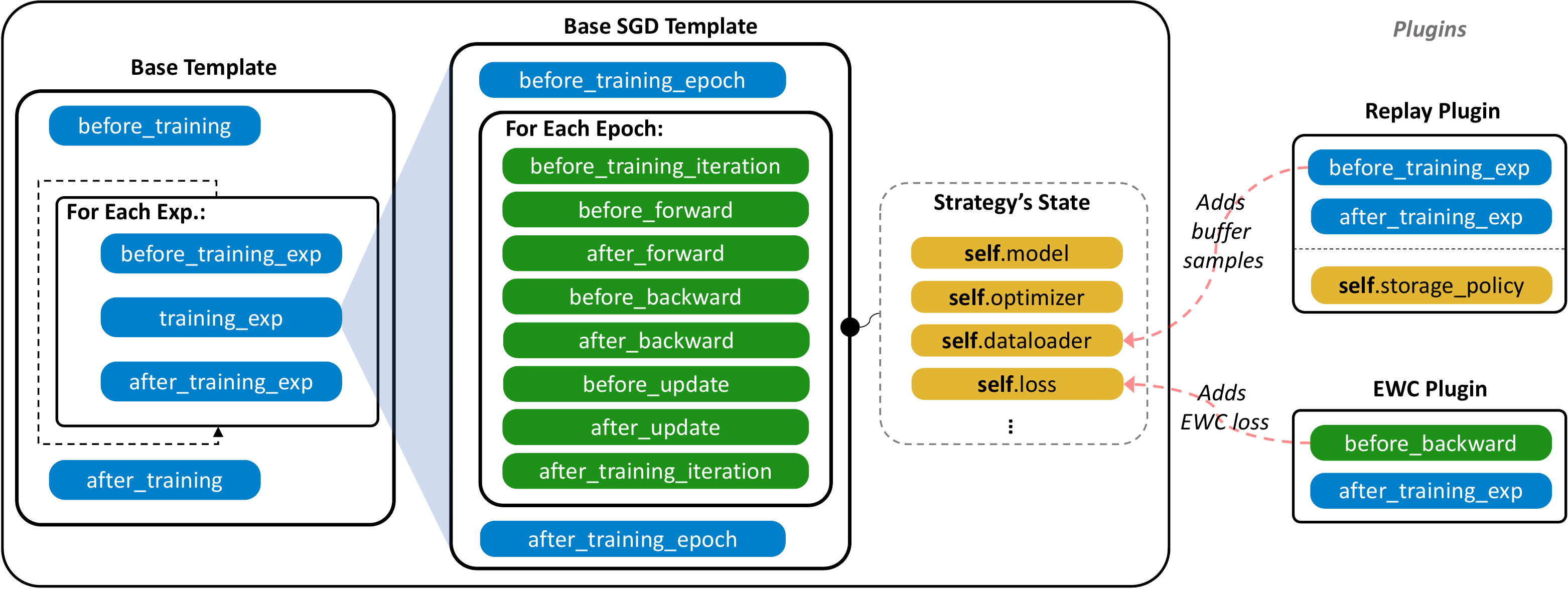}
    \caption{Block diagram of an SGD-based strategy. Replay plugin augments strategy's dataloader while EWC adds a reg. term to the strategy's loss before each update.}
    \label{fig:avl_strategy}
\end{figure}
\paragraph{Inside an Avalanche Strategy} Figure \ref{fig:avl_strategy} shows a high-level overview of an Avalanche strategy. \emph{Templates} define the structure of the training and evaluation loops, providing a set of callbacks that can be used to execute code at any point of the loop. Strategies combine templates to define a complex training loop (e.g., base, SGD, online, supervised, meta-learning). \emph{Plugins} use callbacks to access the strategy state and execute code at specific moments of the loop. Finally, a \emph{Strategy} is an implementation of a set of templates with a list of plugins (e.g., Naive finetuning with ReplayPlugin, EWCPlugin, EvaluationPlugin). Most features in Avalanche can be used as a plugin: training utilities, CL methods, the evaluation and logging system. The advantage of this approach is that any plugin can be used with any template that supports its required callbacks and attributes, making it easy to write general components that can be reused across many experimental settings.

\paragraph{Testing} Avalanche is thoroughly tested with a battery of unit tests. Each pull request is tested by the continuous integration pipeline. A subset of \emph{continual-learning-baselines} (\href{github.com/continualAI/continual-learning-baselines}{link}) is executed with a regular cadence to ensure that Avalanche baselines are in line with expected results from the literature.

\section{Conclusion}

Currently, Avalanche \texttt{v0.3.1} constitutes the largest software library for deep continual learning. Its main focus on \emph{fast prototyping}, \emph{re-producibility} and \emph{portability} makes it the perfect candidate for research-oriented projects. The library is a result of more than two years of development effort involving more than fourteen different research organizations across the world.
The MIT licensed software and the support of \emph{ContinualAI} ensure continuity and alignment with the continual learning research community at large. In the future, we plan to increase the number of available benchmarks and methods, keeping a strong focus on reproducibility and \emph{continual-learning-baselines}(\href{https://github.com/continualAI/continual-learning-baselines}{link}). We also plan to provide better support for emph{reinforcement learning}(\href{https://github.com/ContinualAI/avalanche-rl}{link}), \emph{distributed}, and \emph{federated training} support, while bringing the toolkit to maturity and its first stable official release \texttt{v1.0.0}.



\bibliography{sample}

\end{document}